\documentclass[letterpaper, 10pt, conference]{ieeeconf}

\usepackage{graphicx}
\usepackage{cite}
\usepackage{microtype}

\graphicspath{{figs/}}

\newcommand{\ignore}[1]{%
}

\IEEEoverridecommandlockouts 

\title{Evaluation of a Motion Measurement System for PET Imaging Studies}

\author{Junxiang Wang$^{1}$, Ti Wu$^{2}$, Iulian Iordachita$^{1}$, Peter Kazanzides$^{2}$
\thanks{$^{1}$Dept. of Mechanical Engineering, Johns Hopkins University, Baltimore, MD 21218, USA {\tt\small (email: [jwang334, iordachita]@jhu.edu)}}%
\thanks{$^{2}$Dept. of Computer Science, Johns Hopkins University, Baltimore, MD 21218, USA {\tt\small (email: pkaz@jhu.edu)}}%
}

\begin{document}

\maketitle

\begin{abstract}
  Positron Emission Tomography (PET) enables functional imaging of deep brain structures,
  but the bulk and weight of current systems preclude their use during many natural human activities,
  such as locomotion.
  The proposed long-term solution is to construct a robotic system that can support an imaging system
  surrounding the subject's head, and then move the system to accommodate natural motion.
  This requires a system to measure the motion of the head with respect to the
  imaging ring, for use by both the robotic system and the image reconstruction software.
  We report here the design and experimental evaluation of a parallel string encoder mechanism
  for sensing this motion. Our preliminary results indicate that the measurement system may achieve
  accuracy within 0.5\,mm, especially for small motions, with improved accuracy possible through
  kinematic calibration.
\end{abstract}

\section{Introduction}

\begin{figure}[tbh]
  \centering
  \includegraphics[width=0.75\linewidth]{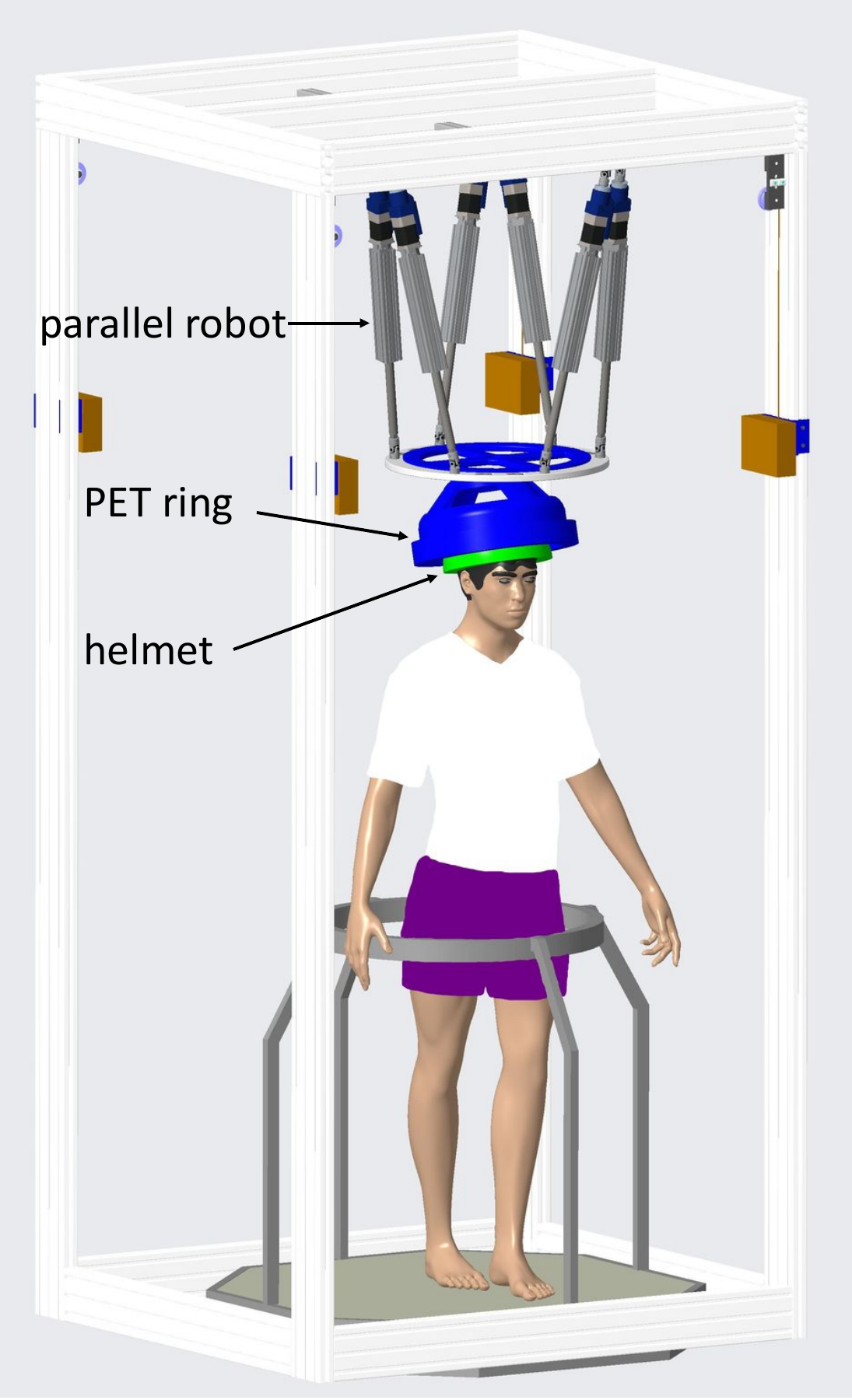}
  \caption{System concept: Parallel robot supports PET imaging ring around a subject's head, while
    the subject walks on a treadmill. The string encoder system measures the position of a helmet (green),
    worn by the subject, with respect to the imaging ring (blue).}
  \label{fig:system-concept}
\end{figure}

Positron Emission Tomography (PET) relies on the injection of a radioactive tracer, which is then
preferentially absorbed by specific tissues (based on the choice of tracer). The absorbed tracer emits
positrons that react with nearby electrons, creating a pair of annihilation (gamma) photons that travel
in opposite directions and are detected by the PET imaging ring.
Higher sensitivity can be achieved by placing the PET detectors as close as possible to the subject.
For brain imaging during natural activities, such as locomotion, the ideal solution would be a wearable
PET imaging ring, such as the Helmet PET \cite{Majewski2011}, except that with current technology,
a PET imaging ring with sufficient sensitivity for neuroscience research would weigh 10\,kg or more
\cite{Majewski2020} (other estimates are 15-20\,kg).
Thus, we are instead exploring the use of a robotic system to suspend the PET imaging ring over
the subject's head and to actively compensate for head motion while the subject performs activities
such as walking on a treadmill, as illustrated in Fig.~\ref{fig:system-concept}.
The goal of the robotic system is to keep the imaging ring approximately
centered around the subject's head (coarse motion compensation), with residual motion corrected
by the image reconstruction algorithm (fine motion compensation).
This is an example of human-robot interaction, where the robot must safely move in proximity of the human.

Given the high safety requirement when robotically moving a 15-20\,kg weight over
a human head, we plan to have a redundant measurement system consisting of both optical and mechanical
sensing of the subject's head motion with respect to the PET imaging ring, possibly fused with inertial sensing.
This paper focuses on the design and experimental evaluation of a mechanical sensing system,
consisting of six string encoders connected between the PET imaging ring and a safety helmet attached
to the subject's head. Our current plan is to use this system as the primary measurement for control
of the robotic system (coarse motion compensation) because it is robust and can provide high-frequency measurements
(on the order of 1\,kHz). In contrast, optical sensing can provide high accuracy, but provides lower-frequency feedback
(on the order of 30-60 Hz) and can occasionally fail to provide a measurement.
We have not yet determined which measurement system to use for the fine motion compensation performed by the
image reconstruction software. According to the imaging scientists, motion measurements should be accurate
to within about 0.5\,mm for this task. Thus, one of our goals is to evaluate whether the mechanical
sensing system can meet this accuracy requirement.

Motion correction is also relevant for conventional PET imaging, and several researchers have investigated
different approaches for sensing this motion.
Especially, different markerless approaches have been studied extensively in recent years.
Olesen \cite{Olesen2011, Olesen2013} describes a system that incorporates a near infrared light emitting diode
into a digital light processing projector to result in a surface scanner, tracking the head motion by creating
a 3D point cloud on the head surface. This system is mounted directly on the PET imaging device, and while
providing high accuracy of less than 0.3\,mm, adds considerable weight to the imaging ring.
This setup also requires clearance between the patient's head and the imaging ring for the emission and
processing of the near infrared light, which is in conflict with our desire for a compact, highly-sensitive
imaging system. Other methods rely more on image processing, thus reducing the amount of computation required
compared to such a 3D surface imaging system. Kyme \cite{Kyme2018} and Anishchenko \cite{Anishchenko2015}
utilized four cameras to record the patient's head, followed by detection of facial features from the video,
and determine the 6 degree-of-freedom (DOF) movement of the head. However, this markerless approach
could only achieve an accuracy of about 2\,mm.

Chamberland \cite{Chamberland2011} uses positron emission fiducial markers to detect tumor movement
for more accurate targeting in radiotherapy. The advantage of this approach is that motion can be directly
measured in the images, which is ideal for the fine motion correction, but is too slow to
control the robot for the coarse motion correction. Also, the fidicual marker would be visible in the
reconstructed image and could interfere with measurements of nearby brain structures.

We previously analyzed motion capture and accelerometer data from subjects during overground and treadmill walking,
respectively, to determine the typical range of head motion, as well as head velocity and acceleration \cite{Liu2021}.
We assumed a helmet of dimension 263\,mm x 215\,mm and evaluated the ability of a simulated
robot to compensate for the recorded head motion, without colliding with an imaging ring of 300\,mm diameter.
Our results indicated that this was feasible, given a robot that could compensate for typical head motions
within about 100\,ms.
Our goal in this work is to develop a measurement system that will enable us to experimentally verify this
assertion. Section \ref{sec:methods} presents the measurement system design and implementation, including integration
of a robot for providing ground-truth displacements. This is followed by accuracy results in Section \ref{sec:results}
and conclusions in Section \ref{sec:conclusions}.

\section{Methods}
\label{sec:methods}

This section describes the design of our mechanical measurement system, using parallel string encoders, followed
by a kinematic analysis, details of the system construction, and homing procedure (to determine initial string
lengths). We next discuss how the string lengths are adjusted to compensate for lateral motion within
their guide channels. Finally, we describe the method for determining the transformation between the UR3 robot
coordinate system and string coordinate system.

\subsection{Mechanical design}
\label{sec:design}

For our mechanical measurement system, we decided to use string encoders, attached between the imaging ring and the helmet.
This creates a parallel structure (essentially a passive parallel robot), where we wish to compute the forward kinematics
to convert the measured joint positions (string encoder lengths) to the Cartesian pose.
Given that we need at least 6 string encoders to provide 6 DOF, there are many different ways that
these can be configured.
We selected a Stewart platform structure over other commonly used cable-driven parallel robots (CDPRs) \cite{Lau2016}
and parallel measuring structures \cite{Jeong1998}, because it is a compact and widely used mechanical
structure \cite{Merlet2006} and is less susceptible to string interference (i.e., the strings should not come into contact
with each other, the imaging ring, or the helmet).

\begin{figure}[tbh]
  \centering
  \includegraphics[width=0.9\linewidth]{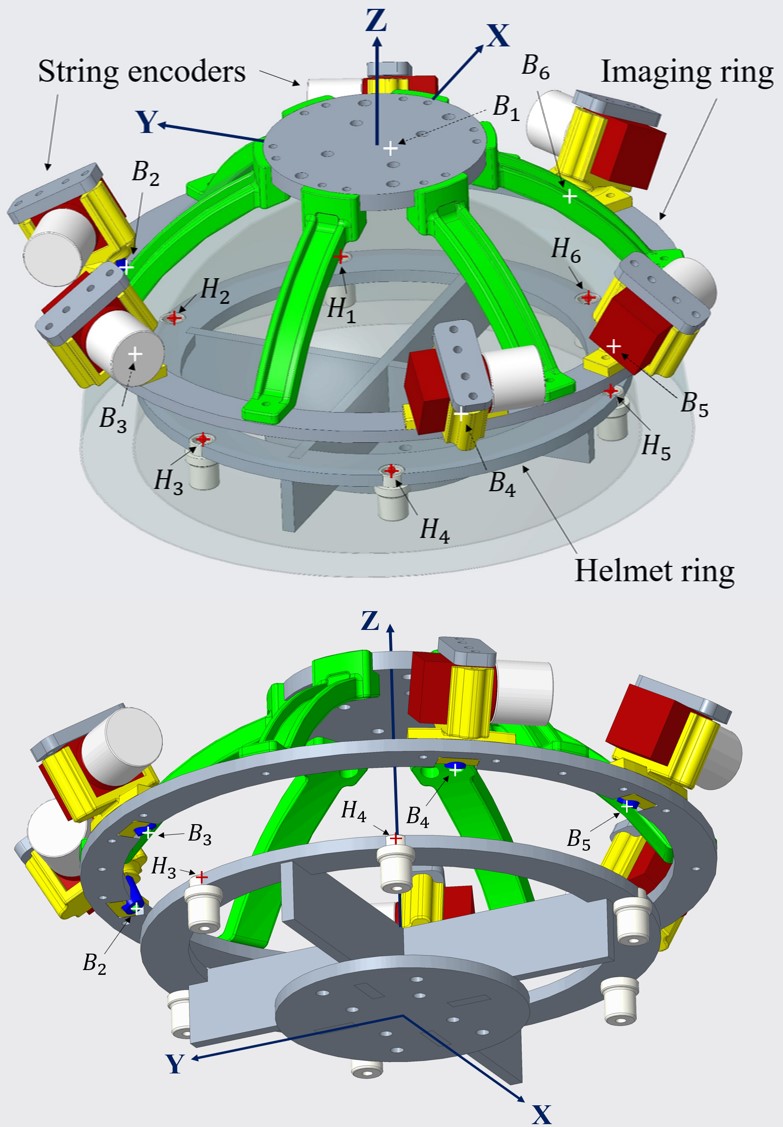}
  \caption{CAD model of string encoder system, with base (imaging ring) and helmet attachment points labeled}
  \label{fig:cad-model}
\end{figure}

We explored several Stewart platform designs, considering motion sensitivity (resolution) and isotropic performance. Details of
the design optimization will published separately.
The final design is based on approximating the imaging ring as a sphere of 300\,mm diameter and the helmet/head as a sphere of
250\,mm diameter. We note that this is just an approximation for our initial evaluation, as the imaging ring is expected to
be cylindrical, not spherical, and the helmet is expected to be elliptical with a major axis of about 263\,mm in diameter.
We created two attachment rings -- one to represent the imaging ring and the other to represent the helmet.
Due to the placement of the rings within the respective spheres, the imaging ring diameter is 261.32\,mm and the helmet
ring diameter is 235.42\,mm. 
Figure~\ref{fig:cad-model} shows a CAD model of the string encoder system generated with Creo. where the top figure indicates
the locations of the string attachment points, and the bottom figure provides a more direct view of the attachment points on the
imaging ring. The orientation of the system is also shown in the figure, and in the nominal configuration, the helmet coordinate
axes are aligned with the base (PET imaging ring) coordinate axes. 
The string attachment points are given in Table \ref{tab:attachment-points}, where $B_i$ and $H_i$ ($i=1\dots6$) are the base and
helmet attachment points respectively, as seen in the nominal configuration.

\begin{table}
  \centering
  \caption{Stewart platform attachment point coordinates (units: mm)}
  \setlength\tabcolsep{4.5pt}
  \begin{tabular}{|c|r|r|r|c|r|r|r|}
    \hline
    \textbf{Point} & \multicolumn{1}{c|}{\textbf{X}} & \multicolumn{1}{c|}{\textbf{Y}} & \multicolumn{1}{c|}{\textbf{Z}} &
    \textbf{Point} & \multicolumn{1}{c|}{\textbf{X}} & \multicolumn{1}{c|}{\textbf{Y}} & \multicolumn{1}{c|}{\textbf{Z}}   \\
    \hline
    $B_1$          & 121.39                          & 48.35                           & 73.67                           &
    $H_1$          & 96.99                           & 66.70                           & 42.06                             \\
    \hline
    $B_2$          & -18.82                          & 129.30                          & 73.67                           &
    $H_2$          & 9.27                            & 117.35                          & 42.06                             \\
    \hline
    $B_3$          & -102.56                         & 80.95                           & 73.67                           &
    $H_3$          & -106.26                         & 50.64                           & 42.06                             \\
    \hline
    $B_4$          & -102.56                         & -80.95                          & 73.67                           &
    $H_4$          & -106.26                         & -50.64                          & 42.06                             \\
    \hline
    $B_5$          & -18.82                          & -129.30                         & 73.67                           &
    $H_5$          & 9.27                            & -117.35                         & 42.06                             \\
    \hline
    $B_6$          & 121.39                          & -48.35                          & 73.67                           &
    $H_6$          & 96.99                           & -66.70                          & 42.06                             \\
    \hline
  \end{tabular}
  \label{tab:attachment-points}
\end{table}

\subsection{String encoder kinematics}

The string encoder system is a Stewart platform and thus we use a standard kinematics formulation,
where the inverse kinematics (from Cartesian pose to string lengths) is easily computed given
the mounting points of the strings on the base (PET imaging ring) and moving platform (helmet).
Specifically, if $B_i$ and $H_i$ ($i=1..6$) represent the coordinates of the base and helmet attachment
points, respectively, and $X$ is the Cartesian pose (transform) of the helmet with respect to the base,
the computed string lengths, $\hat{L}_i$, are given by the following inverse kinematics solution:

\begin{equation}
  \label{eq:fwdkin}
  \hat{L}_i = \left | \left | X*H_i - B_i \right | \right |
\end{equation}

The forward kinematics is more complex and is computed numerically, as described in \cite{Harib2003}.
Given measured string lengths,
$L_i$, the estimated Cartesian pose of the helmet with respect to the base, $\hat{X}$, is
computed iteratively by:

\begin{equation}
  \label{eq:invkin}
  \hat{X}_{k+1} = \hat{X}_k + J(\hat{L}_k) \left ( L_m - \hat{L}_k \right )
\end{equation}

where $k$ is the iteration counter, $J(\hat{L}_k)$ is the Jacobian and $\hat{L}_k$ is the inverse kinematic
solution, eq. (\ref{eq:fwdkin}), corresponding to $\hat{X}_k$.
In this equation, the Cartesian pose is represented by Euler angles, using the intrinsic ZYX
convention, i.e., with respect to the helmet coordinate axes shown in Fig~\ref{fig:cad-model}.
We subsequently refer to the rotation
angle about the z-axis as roll, the angle about the y-axis as pitch, and the angle about the x-axis as yaw.
The iteration is terminated when
the string length error, given by $|| L_m - \hat{L}_k ||$ is less than a specified threshold
(0.01\,mm in our case) or if the iteration counter reaches a specified limit (50 in our
implementation). We typically see convergence in 3-4 iterations.
Note that for a parallel structure, it is more practical to compute the inverse Jacobian and then
obtain the Jacobian by numerically computing the pseudo-inverse.

\subsection{System construction}

\begin{figure}[tbh]
  \centering
  \includegraphics[width=\linewidth]{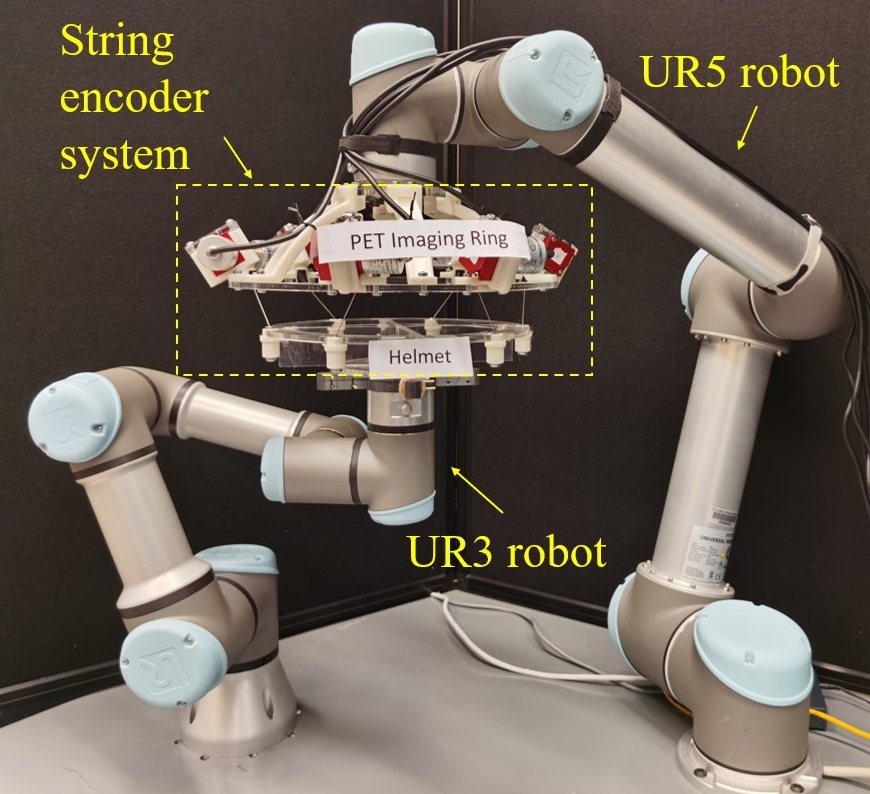}
  \caption{Experimental setup consisting of string encoder system and UR3 and UR5 robots}
  \label{fig:setup}
\end{figure}

To study the motion of the helmet (head) relative to the PET imaging system, using the above presented
kinematic model, we employed an experimental setup consisting of a string encoder measurement system,
with a UR3 robot (Universal Robots, Odense, Denmark) to move the helmet ring and a UR5 robot to
support the PET imaging ring, as shown in Fig.~\ref{fig:setup}. In the experiments reported here,
the UR5 robot is not moved. In future proof-of-concept work, it will move based on the measured head motion to attempt
to keep the (mock) PET imaging ring centered over the helmet.
In the longer term, the weight of an actual PET imaging ring exceeds the UR5 payload and therefore a custom
robot, as shown in Fig.~\ref{fig:system-concept}, will instead be used.
We interfaced to the UR3 robot via TCP, using its real-time script interface.

The six string encoders (MPS-XXXS-200MM-P, Miran Industries, China) each have a measurement range of
200\,mm at a resolution of 60 counts/mm.
The string encoder bodies were attached on the PET imaging ring while the mobile
ends were attached to the helmet. All custom-made parts of the string encoder measurement system were
built with laser-cut acrylic plates combined with 3D printed (Stratasys F170) ABS components. The
geometry of the imaging ring, helmet, and strings connection points were defined based on the Stewart
platform kinematics presented above.

The string encoders are incremental encoders with quadrature outputs (A and B channels) and
an index pulse (I). We created a custom board to interface these signals to an FPGA board developed
for the da Vinci Research Kit (dVRK) \cite{Kazanzides2014}. The FPGA firmware already included
a 4-channel encoder interface module, which was modified to handle up to 8 channels.
We implemented our test software on a PC and connected to the FPGA board via a UDP socket.

\subsection{Homing procedure}

Because the string encoders measure relative displacements, it is necessary to perform a homing
procedure to set the initial absolute displacement. This homing procedure
is performed by utilizing the index pulses provided by the string encoders, which are transmitted on
a different channel than the regular quadrature pulses. Through its 200\,mm range of travel, each string encoder
produces three index pulses, equally spaced apart by 4000 counts, which is equivalent to one-third of the total
measurement range. The location of the first index pulse differs for each string encoder, and the absolute displacement
that corresponds to this first index pulse was recorded before installing each of the string encoders.

With this information, each time the string encoder system is powered on, the measuring ends can be
brought as close to the encoder bodies as possible in order to trigger the first index pulse, which is
captured by the FPGA firmware. By calculating the difference between these values and the known positions,
we can set the offsets to obtain the correct absolute displacements.

\subsection{Calibration of string length adjustment}

\begin{figure}[tbh]
  \centering
  \includegraphics[width=0.9\linewidth]{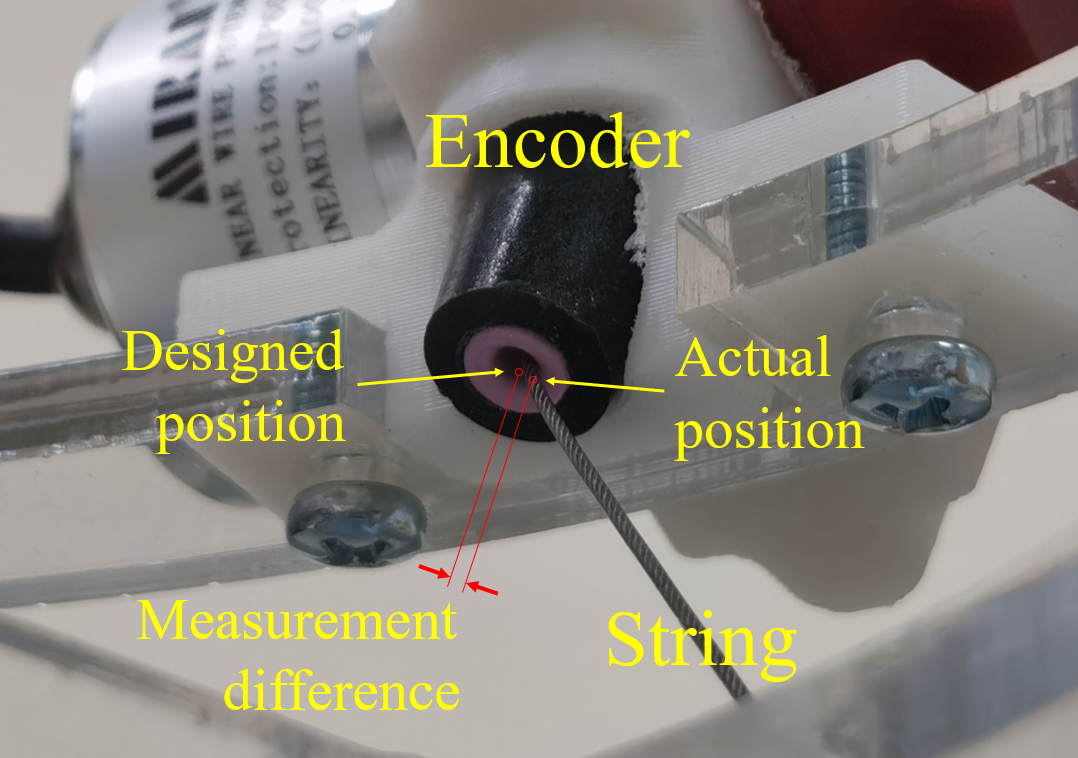}
  \caption{Deviation of the encoder string from the hole center}
  \label{fig:string-hole}
\end{figure}

We observed that the attachment points for the string encoders are affected by a slight deviation from
the ideal model presented earlier. As demonstrated in Fig.~\ref{fig:string-hole}, the designed
attachment point for the string is the center of its circular guide channel on the body of the encoder,
but the string in fact makes contact with the edge of the channel instead of going through the center.
A potential solution would be to adjust the attachment point coordinates to account for the
shift. However, when the helmet platform moves, the string also frequently shifts its position within the
channel, especially when the movement is around the center point. This would require the attachment point
coordinates to be updated during operation, which would complicate the approach.

Instead, we recognized that due to the angled orientation of each string encoder, the string
almost always lies against the edge of the channel---in fact, with our range of motion implemented
below, the string never extends from the center
along the direction of the guide channel. Therefore, the actual reading of each string encoder is less
than the nominal situation where the string passes through the center of the attachment point, and this difference is
always a fixed amount. We then attempted to identify an optimal value for this string length offset
through empirical testing, experimenting with offset lengths ranging from -2\,mm to 6\,mm. For each
setting, we put the robot through a set range of motion in all 6 degrees of freedom and collected a
series of 66 data points. Evaluation of the results and determination of the optimal string length
offset are presented in the following section.

\subsection{Transformation between robot and string encoder system}
\label{sec:transform}

To compare the displacement measured by the string encoder system with the ground-truth displacement
provided by the UR3 robot, a transformation between the two coordinate systems is required.
We accomplished this by considering that the transformation between the PET imaging ring (base) and the
robot, $^BE_R$, is fixed for this experiment because the UR5 robot does not move.
The string encoder system measures the transformation between the PET imaging ring and
the helmet, $^BE_H$, and after setting the robot tool center point (TCP) to be coincident with the center
of the helmet, the robot interface can output the corresponding transformation between the robot and the
helmet, $^HE_R$. Consequently, $^BE_R = {^BE_H} {^HE_R}$, and for a constant position of the base,
the string encoder system only needs to query the robot once to obtain $^BE_H$ and then
compute $^BE_R$. Afterwards, the measurements from the string encoder system can be converted to a 6 DOF
position in the robot coordinate system by computing ${(^BE_H)^{-1}}{^BE_R}$, and the full 6 DOF pose
can be extracted from the resulting transformation.

\section{Results}
\label{sec:results}

\subsection{Experimental setup}

\begin{figure}[tbh]
  \centering
  \includegraphics[width=0.9\linewidth]{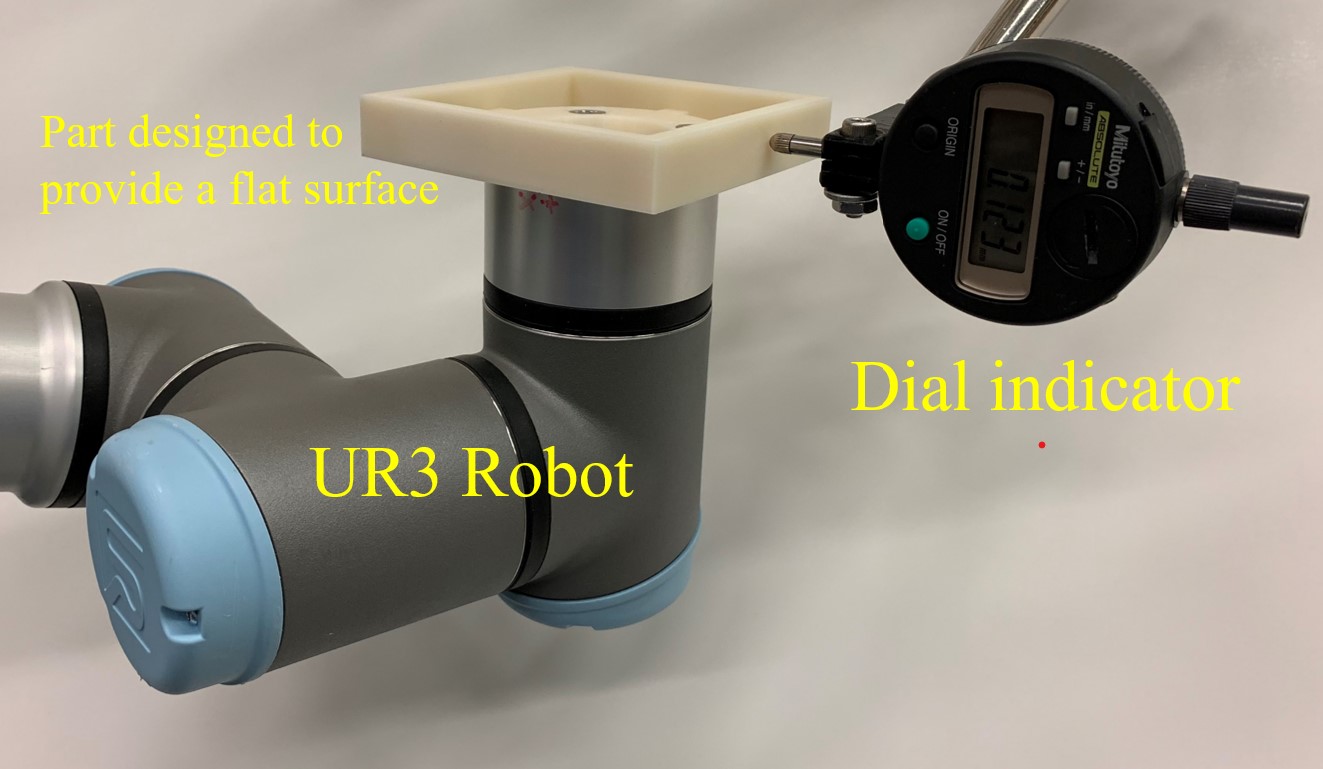}
  \caption{Dial indicator for measuring robot translation accuracy}
  \label{fig:dial-indicator}
\end{figure}

We used a UR3 robot to provide precise motions of the helmet. We first verified the measurement
accuracy of the UR3 robot, for a $\pm$10\,mm range of motion, using a dial indicator (543-693B,
Mitutoyo Corp., Japan), as shown in Fig.~\ref{fig:dial-indicator}. We limit the robot motion to
this specific range since the designed clearance between the helmet and the imaging ring is only
about 15\,mm, as outlined in Section \ref{sec:design}. Table \ref{tab:dial-indicator}
summarizes the result of the verification. The difference indicated in the third column could
have been partially caused by the rough surface on the 3D-printed part that was mounted on the UR3
end-effector.
Nevertheless, we can conclude that the UR3 accuracy is about 0.1\,mm, and hence precise
enough to provide the ground-truth helmet motion in our experiments.

\begin{table}
  \centering
  \caption{UR3 robot accuracy evaluation, showing commanded robot displacement, measured dial indicator displacement
    and difference (units: mm)}
  \begin{tabular}{|r|r|r|}
    \hline
    \textbf{Robot} & \textbf{Dial Indicator} & \textbf{Difference} \\
    \hline
    -10            & -9.880                  & -0.120              \\
    \hline
    -9             & -8.884                  & -0.116              \\
    \hline
    -8             & -7.898                  & -0.102              \\
    \hline
    -7             & -6.912                  & -0.088              \\
    \hline
    -6             & -5.911                  & -0.089              \\
    \hline
    -5             & -4.914                  & -0.086              \\
    \hline
    -4             & -3.950                  & -0.050              \\
    \hline
    -3             & -2.951                  & -0.049              \\
    \hline
    -2             & -1.946                  & -0.054              \\
    \hline
    -1             & -0.956                  & -0.044              \\
    \hline
    0              & 0.004                   & -0.004              \\
    \hline
    1              & 0.966                   & 0.034               \\
    \hline
    2              & 1.956                   & 0.044               \\
    \hline
    3              & 2.956                   & 0.044               \\
    \hline
    4              & 3.923                   & 0.077               \\
    \hline
    5              & 4.925                   & 0.075               \\
    \hline
    6              & 5.909                   & 0.091               \\
    \hline
    7              & 6.913                   & 0.087               \\
    \hline
    8              & 7.899                   & 0.101               \\
    \hline
    9              & 8.881                   & 0.119               \\
    \hline
    10             & 9.874                   & 0.126               \\
    \hline
  \end{tabular}
  \label{tab:dial-indicator}
\end{table}

\subsection{Calibration results}

\begin{figure}[tbh]
  \centering
  \includegraphics[width=1.0\linewidth]{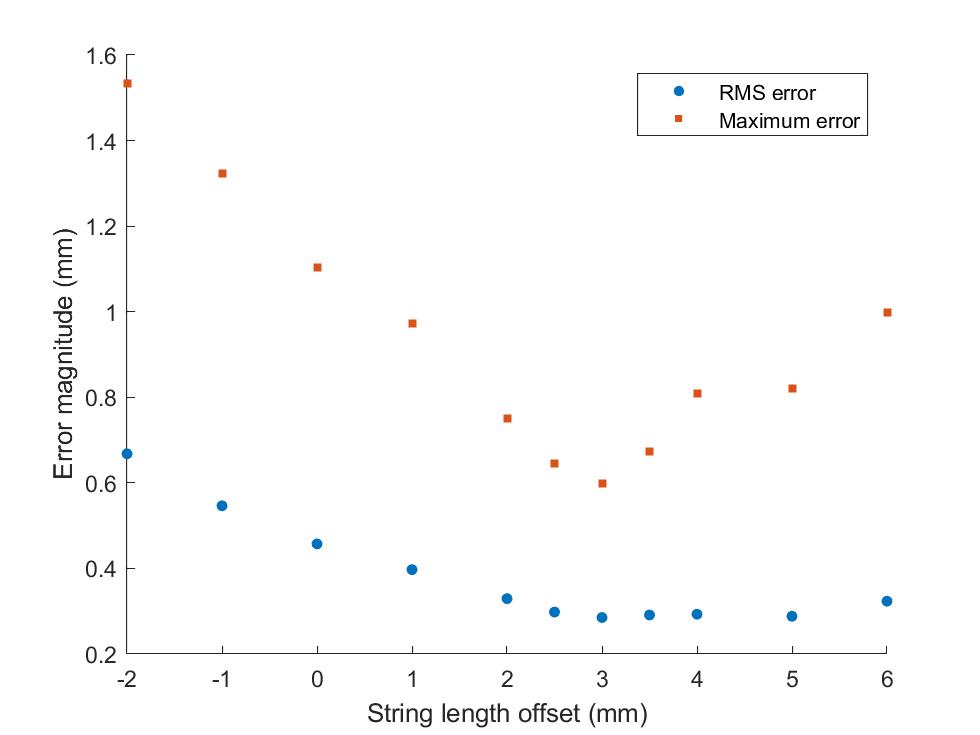}
  \caption{Calibration result for string length offsets}
  \label{fig:calibration}
\end{figure}

For each string length offset, we obtained the vector difference $\Delta P$ between the displacement
measured by the string encoder system, $P_{enc}$, and the actual displacement produced by the UR3
robot, $P_{rob}$, at each of the 66 points. We then took the norm of these error vectors, $|\Delta P|$,
obtained the root-mean-square (RMS) as well as the maximum across all 66 points, and plotted them against the respective
offset, as shown in Fig.~\ref{fig:calibration}. All offsets from 2.5\,mm to 5\,mm share similar RMS
error magnitudes of around 0.3\,mm, but the 3\,mm offset is shown to have the smallest maximum error
magnitude.

We also used calipers to measure the string outer diameter, which was 0.8\,mm, and the inner diameter of its guide channel,
which was 2.8\,mm, giving a clearance of 1\,mm.
A similar phenomenon is also present on the moving end of the encoder attached on the helmet, where the
position of the string is also shifted by about 1\,mm. Additionally, we observed the difference in the
encoder readings before and after temporarily holding one string at the two designated attachment points
using tools, and noted that the string length reading increased by 2.5\,mm. Therefore, our selection
of the 3\,mm offset based on error minimization is supported by physical measurement, and this value
is used in the subsequent accuracy evaluation.

\subsection{Accuracy evaluation}

\begin{figure}[tbh]
  \centering
  \includegraphics[width=1.0\linewidth]{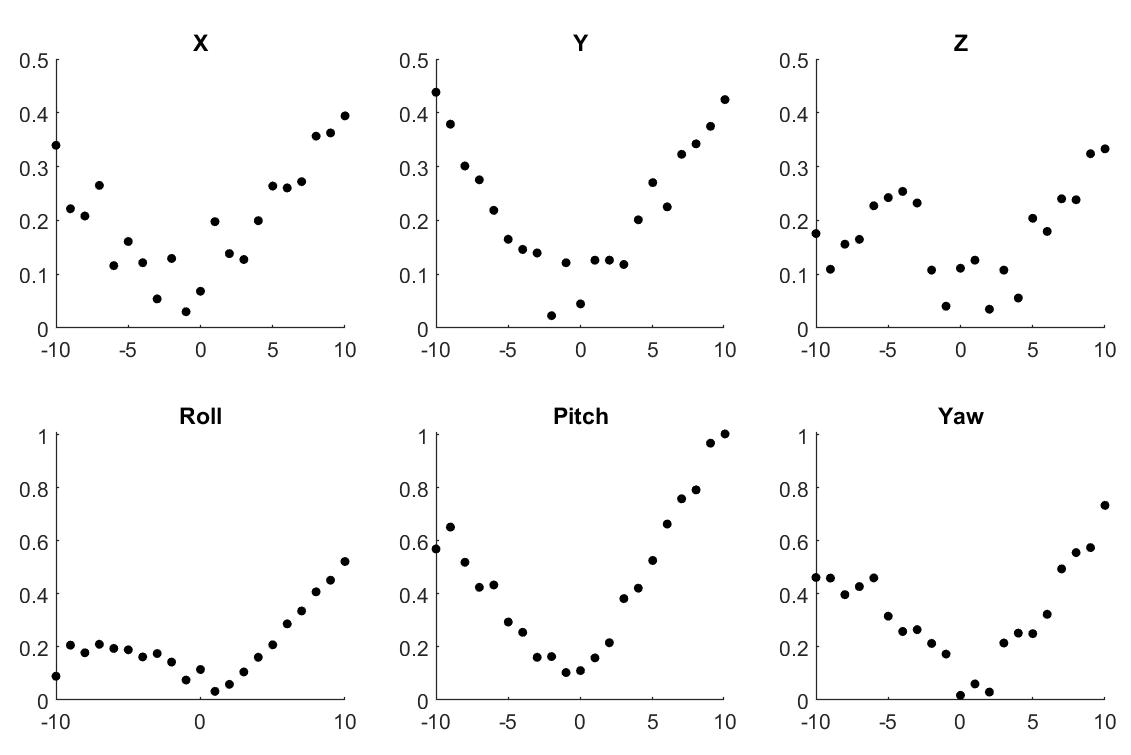}
  \caption{Accuracy evaluation result of the string encoder system.
    Horizontal axes represent displacements (mm or deg), and vertical axes represent the error magnitude $|\Delta P|$ (mm)}
  \label{fig:accuracy}
\end{figure}

We first determined the transformation between the UR3 robot coordinate system and the string encoder coordinate system,
as described in Section \ref{sec:transform}.
Then, the UR3 robot performed a range of motion of
$\pm$10\,mm or deg along each of the axes, one axis at a time, in increments of 1\,mm or deg.
At each position, the string lengths were recorded, adjusted by the 3\,mm string length offset identified above,
and then the forward kinematics applied to compute the Cartesian position.
The results are shown in Table \ref{tab:accuracy} as well as in Fig.~\ref{fig:accuracy}.

\begin{table}
  \centering
  \caption{Translation error, $|\Delta P|$ (mm), due to displacements (in mm or deg) from origin,
    $\Delta P = P_{enc} - P_{rob}$.}
  \begin{tabular}{||c||c|c|c|c|c|c||}
    \hline
    \textbf{Displacement} & \multicolumn{6}{c||}{\textbf{$|\Delta P|$ (mm)}} \\
    \cline{2-7}
    \textbf{(mm or deg)}  & \textbf{X}     & \textbf{Y}     & \textbf{Z}     & \textbf{Roll}  & \textbf{Pitch} & \textbf{Yaw}   \\
    \hline
    -10                   & 0.340          & 0.438          & 0.176          & 0.090          & 0.569          & 0.462          \\
    \hline
    -9                    & 0.222          & 0.379          & 0.109          & 0.207          & 0.652          & 0.460          \\
    \hline
    -8                    & 0.208          & 0.301          & 0.156          & 0.178          & 0.519          & 0.397          \\
    \hline
    -7                    & 0.265          & 0.275          & 0.165          & 0.210          & 0.425          & 0.428          \\
    \hline
    -6                    & 0.116          & 0.219          & 0.227          & 0.194          & 0.433          & 0.460          \\
    \hline
    -5                    & 0.161          & 0.165          & 0.242          & 0.189          & 0.294          & 0.316          \\
    \hline
    -4                    & 0.121          & 0.146          & 0.254          & 0.162          & 0.255          & 0.258          \\
    \hline
    -3                    & 0.054          & 0.139          & 0.232          & 0.175          & 0.161          & 0.265          \\
    \hline
    -2                    & 0.129          & 0.023          & 0.108          & 0.143          & 0.164          & 0.213          \\
    \hline
    -1                    & 0.030          & 0.121          & 0.040          & 0.075          & 0.103          & 0.173          \\
    \hline
    0                     & 0.068          & 0.045          & 0.111          & 0.115          & 0.111          & 0.018          \\
    \hline
    1                     & 0.198          & 0.126          & 0.126          & 0.033          & 0.159          & 0.061          \\
    \hline
    2                     & 0.138          & 0.126          & 0.035          & 0.059          & 0.216          & 0.030          \\
    \hline
    3                     & 0.127          & 0.118          & 0.108          & 0.106          & 0.382          & 0.215          \\
    \hline
    4                     & 0.199          & 0.201          & 0.056          & 0.161          & 0.421          & 0.252          \\
    \hline
    5                     & 0.264          & 0.270          & 0.204          & 0.208          & 0.526          & 0.250          \\
    \hline
    6                     & 0.260          & 0.225          & 0.179          & 0.287          & 0.663          & 0.323          \\
    \hline
    7                     & 0.272          & 0.323          & 0.240          & 0.336          & 0.758          & 0.494          \\
    \hline
    8                     & 0.357          & 0.342          & 0.238          & 0.408          & 0.792          & 0.556          \\
    \hline
    9                     & 0.363          & 0.375          & 0.324          & 0.451          & 0.968          & 0.574          \\
    \hline
    10                    & 0.394          & 0.425          & 0.333          & 0.522          & 1.003          & 0.734          \\
    \hline
    \textbf{RMS}          & \textbf{0.223} & \textbf{0.257} & \textbf{0.193} & \textbf{0.241} & \textbf{0.529} & \textbf{0.377} \\
    \hline
  \end{tabular}
  \label{tab:accuracy}
\end{table}

The results indicate that we have achieved RMS errors less than 0.3\,mm for translations along all three axes,
and for rotation around the z-axis, and less than 0.6\,mm for the other two rotations.
For the translations, even the maximum error
magnitude within the $\pm$10\,mm motion range was less than the 0.5\,mm requirement for fine motion correction.

\section{Conclusions}
\label{sec:conclusions}

We developed a mechanical 6 DOF measurement system consisting of 6 parallel string encoders, connected in a Stewart
platform configuration. This system is intended to be used to measure the motion of a helmet, worn by a human subject,
with respect to a PET imaging ring supported by a robotic system. We performed experiments, using a robot to provide
precise helmet motions, and verified that RMS translation errors are usually less than 0.3\,mm, with values slightly above 0.5\,mm
for some rotations. These relatively larger errors associated with displacements in the pitch and yaw angles might be
attributed to inaccuracy in setting the robot TCP, a process mentioned in section \ref{sec:transform}. Since the TCP
is set to be a constant offset in the direction (Z) perpendicular to the mounting surface of the robot end effector, an
inaccuracy in setting this offset would cause robot rotations about the tool X and Y axes (yaw and pitch, respectively)
to also cause translations that would be measured by the string encoder system.
Nonetheless, the performance described above is sufficient for coarse motion correction, where a robot moves the
PET imaging ring to attempt to keep it centered around the head. It may be suitable for fine motion correction, performed
during image reconstruction, where an accuracy of 0.5\,mm or better is desired.

Our testing only evaluated one motion axis at a time---it is possible that combined motions (e.g., simultaneous
translation and rotation) would produce higher errors.
We have not yet identified a specification for the largest rotation error and our testing did not measure the
rotation error (we only measured the translation error resulting from applied rotations).
Considering that the maximum radius of a human head is about 100\,mm, a rotation error of 0.3 degrees would produce
an error of about 0.5\,mm on the skull surface. However, for neuroscience research, targets of interest will likely be
closer to the center of the head and therefore the effect of rotation error may be less critical. Our future work
includes determination of the rotation accuracy specification for fine motion correction.
In addition, if the UR5 robot is able to keep up with the head motion, the relative displacement of the head with respect to the
imaging ring may be much smaller than the 10\,mm and 10 degree displacements tested here. In that case, we could
expect higher accuracy as our results indicate higher accuracy for displacements closer to the center position.
If necessary, we can perform a kinematic calibration of the string encoder system
using methods developed for Stewart platforms \cite{Zhuang1995} to further increase the accuracy.

Our next stage of development will be to use the UR3 robot to emulate realistic human head motion, using the previously
recorded data analyzed in our prior work \cite{Liu2021}, and to use the UR5 robot to move the PET imaging ring based
on those measurements. Eventually, since the PET imaging ring can weigh up to 20\,kg and the UR5 has a payload of only
5\,kg, the UR5 would be replaced by a custom robot, but currently it serves as a prototype for verification of our
motion measurement and compensation system.

\section*{Acknowledgements}
Yangzhe Liu participated in early design discussions of the measurement system.

\bibliographystyle{IEEEtran}
\bibliography{references}
\end{document}